\documentclass[lettersize,journal]{IEEEtran}
\usepackage{amsmath,amsfonts}
\usepackage{algorithmic}
\usepackage{algorithm}
\usepackage{array}
\usepackage[caption=false,font=normalsize,labelfont=sf,textfont=sf]{subfig}
\usepackage{textcomp}
\usepackage{stfloats}
\usepackage{url}
\usepackage{verbatim}
\usepackage{graphicx}
\usepackage{cite}
\usepackage{amsthm,amsmath,amssymb}
\usepackage{mathrsfs}
\usepackage{booktabs,multirow} 
\usepackage{colortbl}  %彩色表格需要加载的宏包
\usepackage{xcolor}
\begin{document}

\title{Camera-aware Label Refinement for \\ Unsupervised Person Re-identification}

\author{Pengna Li, Kangyi Wu, Wenli Huang, Yang Wu, Sanping Zhou, and Jinjun Wang %~\IEEEmembership{Student member,~IEEE}
        % <-this % stops a space
%\thanks{This paper was produced by the IEEE Publication Technology Group. They are in Piscataway, NJ.}% <-this % stops a space
\thanks{The first two authors made equal contributions to the paper writing and experiments. (Corresponding author: Jinjun Wang)}
\thanks{Pengna Li, Kangyi Wu, Yang Wu, Sanping Zhou, Jinjun Wang are with the Institute of Artificial Intelligence and Robotics, Xi’an Jiaotong University, Xi’an, Shaanxi 710049, China.(e-mail:{sauerfisch, wukangyi747600, wuyang\_cc}@stu.xjtu.edu.cn, spzhou@xjtu.edu.cn, jinjun@mail.xjtu.edu.cn)}
\thanks{Wenli Huang is with the School of Electronic and Information Engineering, Ningbo University of Technology, Ningbo, Zhejiang 315211, China.(e-mail:huangwenwenlili@126.com)}}
\markboth{Journal of \LaTeX\ Class Files,~Vol.~14, No.~8, March~2024}%
{Shell \MakeLowercase{\textit{et al.}}: A Sample Article Using IEEEtran.cls for IEEE Journals}


\maketitle

\begin{abstract}
Unsupervised person re-identification aims to retrieve images of a specified person without identity labels. Many recent unsupervised Re-ID approaches adopt clustering-based methods to measure cross-camera feature similarity to roughly divide images into clusters. They ignore the feature distribution discrepancy induced by camera domain gap, resulting in the unavoidable performance degradation. Camera information is usually available, and the feature distribution in the single camera usually focuses more on the appearance of the individual and has less intra-identity variance. Inspired by the observation, we introduce a \textbf{C}amera-\textbf{A}ware \textbf{L}abel \textbf{R}efinement~(CALR) framework that reduces camera discrepancy by clustering intra-camera similarity. Specifically, we employ intra-camera training to obtain reliable local pseudo labels within each camera, and then refine global labels generated by inter-camera clustering and train the discriminative model using more reliable global pseudo labels in a self-paced manner. Meanwhile, we develop a camera-alignment module to align feature distributions under different cameras, which could help deal with the camera variance further. Extensive experiments Market1501, DukeMTMC-reID, MSMT17 and Veri-776 validate the superiority of our proposed method over state-of-the-art approaches. The code is accessible at \url{https://github.com/leeBooMla/CALR}.
\end{abstract}


\section{Introduction}
\IEEEPARstart{P}{erson} re-identification~(Re-ID) is a task to identify a person corresponding to a given query under disjoint cameras.~\cite{ye2021deep}
% It has drawn increasing attention in multimedia and computer vision communities due to its broad research impact and practical application in large-scale intelligent surveillance systems and public safety~\cite{fox2007robust, wang2015zero, ye2016person, zhou2017large, wang2019incremental}. Besides, many multimedia tasks such as multi-object tracking~\cite{wang2020towards}, person activity analysis~\cite{loy2009multi} and visual navigation~\cite{krantz2023navigating} can benefit from the advancements of the Re-ID techniques. 
% The goal of Re-ID is to identify a specified person image under one camera and retrieve the matching images of the person under other cameras. 
% It is a challenging task due to the intensive variations in illumination, camera viewpoints and resolutions, occlusion, pedestrian poses and cloths, etc. These challenges impede the performance of Re-ID model. In recent years, many methods have been proposed to address this challenging task. 
With the advancement of deep learning, supervised Re-ID methods~\cite{shu2021large,he2021transreid, he2023fastreid} have achieved significant performance improvement. 
% They utilize personal identity information to enhance the robustness of Re-ID model to varied illumination~\cite{lu2023illumination}, pedestrian pose~\cite{ wang2021horeid} and cloths~\cite{yu2021apparel}, occlusion~\cite{jia2022learning} and camera domain shift~\cite{zhuang2020rethinking}. 
Unfortunately, purely supervised methods heavily rely on a large quantity of expensive annotated data, which limits their adaptability to practical applications. Recently, there has been increasing research focus on unsupervised settings to alleviate the data annotating requirements. 

Unsupervised Re-ID approaches can be broadly divided into unsupervised domain adaption (UDA) approaches and purely unsupervised learning approaches based on whether they use external annotated Re-ID dataset. The UDA line~\cite{li2021adadc,tao2021unsupervised,liu2022complementary,hu2022divide, lee2023cameradriven} has demonstrated notable performance gains with the availability of knowledge from the source domain. However, their performance is contingent upon the quality and reliability of the source domain. 

\begin{figure}[t!]
\centering
\includegraphics[width=8.5cm]{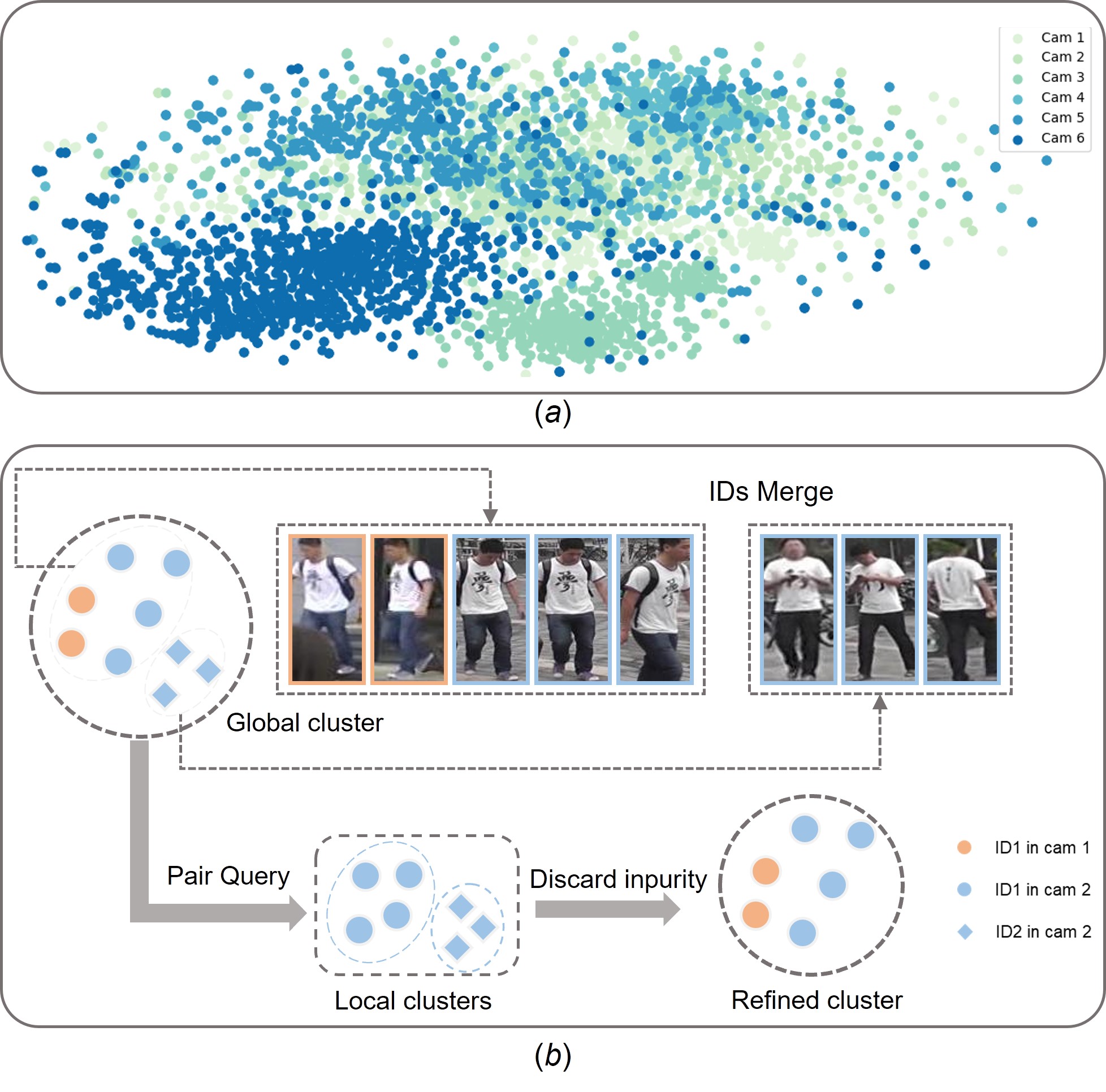}
\caption{We illustrate the T-SNE visualization~\cite{van2008visualizing} in (a) for the feature distribution on Market-1501~\cite{zheng2015scalable}, where features are extracted using ResNet-50 pre-trained on ImageNet\cite{deng2009imagenet}. Each color indicates samples from different cameras. Feature distributions are highly biased towards camera labels. Consequently, positive pairs captured from different cameras may exhibit greater dissimilarity than negative samples from the same camera, resulting in what we refer to as "IDs Merge" as shown in (b). To address this issue, we exploit more fine-grained and reliable local labels generated in advance to refine global clusters.}
\label{img1}
\hspace{0.1in}
\end{figure}

In this paper, we tackle the more challenging yet practical task, purely unsupervised Re-ID, where the model is trained without any identity labels. Most existing unsupervised approaches adopt a certain pseudo-label-based scheme that alternates between assigning similar images with the same pseudo labels via clustering~\cite{lin2019bottom,dai2022cluster}, softened labels~\cite{lin2020unsupervised, wang2020unsupervised} or label estimation~\cite{ji2021meta}, and then the model is trained using these obtained pseudo-labels. Here, we adopt a two-stage clustering-based methods with a simple and flexible pipeline. 

Although clustering-based methods have been attempted intensively~\cite{lin2019bottom, dai2022cluster}, we argue that two main issues impede the performance of existing approaches: 1) The inherent label noise arises from the variations of body pose, background, and camera resolution. Such noise propagates and accumulates during the training process, leading to a degradation in the model's performance; 2) The feature distribution discrepancy across camera domains, which makes it challenging to learn consistent representations for the same ID. In addition, the two issues are interrelated, which further complicates the situation. For the first issue, several studies use local part features~\cite{cho2022part}, group labels~\cite{peng2023adaptive} and other approaches~\cite{li2024distribution,yang2025idenet} to improve the accuracy of pseudo labels. For the second issue, efforts have been made in GAN-based generation~\cite{Zhong_Zheng_Luo_Li_Yang_2019, Lin_Wu_Yan_Xu_Yang_2020}, data alignment~\cite{Zhuang_Wei_Xie_Zhang_Zhang_Wu_Ai_Tian_2020, Yang_Zhong_Luo_Cai_Lin_Li_Sebe_2021} and other techniques~\cite{wang2021camera, zhang2023camera}. However, it is seen that there still exists a large gap compared to supervised approaches.

To overcome the aforementioned problems, we introduce a camera-aware label refinement~(CALR) framework to deal with the label noise and camera discrepancy. Our idea is inspired by the below phenomenon. As illustrated in Fig.~\ref{img1}~(a), image feature distribution suffers from strong camera bias. Due to the contribution of the camera domain, persons captured within the same camera tend to cluster closer than those captured by different cameras. If we roughly measure the feature similarity and divide images into clusters, it would merge images from the same camera but with different IDs into a single cluster, as demonstrated in Fig.~\ref{img1}~(b), leading to the ``IDs Merge" error and irreversible performance drop. Nevertheless, features in a single camera could be free from the influence of camera view and focus more on discriminating the pedestrian appearance. Therefore, we instead conduct intra-camera clustering for each camera to observe local clusters. We notice that even appearance-alike persons with different IDs can be accurately separated. This motivated us to utilize the local clusters to discard impurities in global clusters. To further mitigate the camera discrepancy problem, we introduce a camera-domain alignment module to pursue consistent distribution among different cameras. In this way, the two issues mentioned above can be addressed. Specifically:

1) For the first issue, we refine the global labels with more reliable local ones to cope with the inherent label noise. We employ a two-stage training scheme to optimize the Re-ID model, ~\emph{i.e.}, $a$) \textbf{intra-camera training} conducted within singe camera respectively. Features from each camera are extracted and clustered to generate independent local pseudo-labels. These pseudo-labels serve as supervision to optimize their respective encoders. This step ensures that the local pseudo-labels are sufficiently reliable to refine the global inter-camera clustering results in the next stage; $b$) \textbf{inter-camera training} conducted across cameras. We first select some pivot nodes for each cluster, typically those with high utility. For each pivot, we query the relationships with nodes belonging to the same cluster and eliminate negative samples based on the cluster results obtained in the first stage. Therefore, the remaining samples are more reliable for learning. Besides, as the training process, we progressively decay the probability of discarding samples, which enables us to train the Re-ID model through a self-paced way~\cite{bengio2009curriculum}.
    
2) For the second issue, we develop a camera domain alignment module designed to handle the feature distribution discrepancy and mitigate the influence of camera variance. The idea is realized by domain-adversarial learning to learn better feature representations, which utilize a gradient reversal layer (GRL)~\cite{ganin2016domain} to add a domain classifier to the feature encoder. It ensures the consistency of the features among different cameras. To the best of our knowledge, this is the pioneering effort to employ domain-adversarial learning for aligning feature distributions of different cameras in the purely unsupervised person Re-ID task. 

In our previous work~\cite{li2023pseudo}, we adopted intra-camera clustering results to refine global labels, which does not solve the feature discrepancy explicitly. This paper introduces a new camera domain alignment module and provides a more comprehensive experimental evaluation of three person Re-ID datasets and one vehicle Re-ID dataset.  We summarize the main contributions of our paper as follows:

\begin{itemize}
\item{A novel camera-aware label refinement framework is developed with reliable and fine-grained local labels, which adequately exploits intra-camera similarity to deal with pseudo label noise.}

\item{We define a new centrality criterion to estimate the utility of node and conduct refinement decaying strategy, which helps refine global labels accurately and optimize the Re-ID model in a self-paced manner.}

\item{A camera domain alignment module is proposed to alleviate feature distribution discrepancy caused by the camera bias, which facilitates optimizing the Re-ID model and learning better feature representations.}

\item{Extensive qualitative and quantitative experiments demonstrated our proposed CALR surpasses the state-of-the-art methods on multiple large-scale datasets.}
\end{itemize}

The rest of the paper is organized as follows. Section~\ref{sec2} provide a comprehensive review of the related works. Section~\ref{sec3} details the methodologies. In Section~\ref{sec4}, we present the results and analysis of our experiments. Section~\ref{sec5} concludes the paper, summarizing the key findings and contributions.

\section{Related works} \label{sec2}
\subsection{Learning with Noisy Labels}
In real-world scenarios, collecting high-quality labels is expensive while cheap but noisy labels are more readily available. Hence, learning with noisy labels is becoming increasingly popular. There are numerous approaches to address the task, encompassing techniques such as designing robust architecture~\cite{cheng2022instance}, improving loss function~\cite{zhang2018generalized}, introducing robust regularization~\cite{zhang2017mixup} and selecting confident samples~\cite{huang2019o2u}, \emph{etc}. For the unsupervised person Re-ID task, generated labels are usually noisy in the early training. Yuan~\cite{yuan2020defense} proposes a fast-approximated triplet loss to explicitly handle label noise. Zhao~\cite{li2018unsupervised} develops a noise-resistible mutual-training method to train two networks. In comparison, this paper is motivated to select more confident samples for global clusters based on intra-camera local labels.

\subsection{Unsupervised Person Re-ID}

Unsupervised person Re-ID can be categorized into UDA person Re-ID and purely unsupervised Re-ID based on whether using external annotated data. 

{\bf{UDA person Re-ID}} leverage the annotated data from the source domain to adjust the model to the target domain without requiring any ID information. To address this task, existing methods focus on feature distributions alignment or knowledge transfer between source and target domains to alleviate domain gap and learn domain-invariant representations. Some researchers usually adopt Generative Adversarial Networks (GAN)~\cite{goodfellow2014generative} to perform image-image translation~\cite{zhu2017unpaired}, which transfer the style of the source images to match that of the target domain~\cite{wei2018person, verma2021unsupervised} or restrain the background bias~\cite{huang2019sbsgan} to eliminate the feature distribution discrepancy between different domains. Other researchers employ the knowledge gained from the source domain to cluster unlabeled data for the target domain and iteratively refine the model by generating pseudo labels using a self-training scheme. AD-Cluster~\cite{zhai2020ad} augments person clusters to enhance the discrimination capability of the Re-ID encoder. SPCL~\cite{ge2020self} proposes a unified contrastive learning framework to train the source and target domain jointly. Bai~\cite{bai2021unsupervised} utilizes multiple source datasets to combine more knowledge to adapt the target domain. 
% AWB~\cite{wang2022attentive} develops a mutual learning approach using the dual network to produce reliable soft pseudo labels.
Despite the employment of annotated auxiliary datasets, recent UDA works fail to demonstrate significant advantages over purely unsupervised methods. Different from these methods, our focus is on the purely unsupervised person Re-ID without requiring any identity annotation. 

{\bf{Purely unsupervised person Re-ID}} is to perform person Re-ID solely on the unlabeled target domain, which relies entirely on unsupervised learning methods to identify individuals without any identity labels. Recent works~\cite{dai2022cluster, lin2019bottom} perform cluster algorithms in target features and assign pseudo labels to images. 
BUC~\cite{lin2019bottom} employs a bottom-up approach to progressively cluster similar images into the same classes. 
ClusterContrast~\cite{dai2022cluster}, DCCT~\cite{chen2023dual} and DHCCN~\cite{li2024distribution} 
are proposed to optimize the contrastive learning framework to learn a discriminative model. ClusterContrast~\cite{dai2022cluster} stores and updates cluster representations and computed ClusterNCE loss at the cluster level. DCCT~\cite{chen2023dual} introduces a dual clustering co-teaching method to utilize the features extracted by two networks to generate two sets of pseudo-labels respectively. DHCCN~\cite{li2024distribution} proposes a distribution-guided hierarchical calibration contrastive learning framework and utilize low-confidence samples to correct the features distribution.
% HCM~\cite{si2022hybrid} is designed to explore hard sample feature similarities in the image-level and identity-level. DiDAL~\cite{liu2023discriminative} matches an anchor with multiple hard positive samples to reduce the label noise. HCACE~\cite{luo2024hierarchical} exploits hierarchical camera-aware relations at proxy-level and instance level. CIFL~\cite{pang2022camera} introduce a new DBSCAN-NN clustering algorithm to improve the intra-class camera diversity to enhance the accuracy of labels. 
However, clustering-based methods using hard labels tend to accumulate clustering errors during iterations. To tackle the problem, some researchers~\cite{lin2020unsupervised, wang2020unsupervised} discard clustering and assign unlabeled images with softened multi-class labels reflecting identity. Others~\cite{ge2020mutual, zhang2021refining, cho2022part} attempt to handle the label noise using label refinement methods. MMT~\cite{ge2020mutual} proposes a mutual mean-teaching framework to mitigate label noise. RLCC~\cite{zhang2021refining} delves into temporal cluster consensus to improve the reliability of pseudo labels. PPLR~\cite{cho2022part} utilizes fine-grained local context information and ensembles the prediction of part features to refine pseudo labels of global features. In this paper, we follow the clustering-based methods and propose a camera-aware label refinement framework to enhance the quality of pseudo labels.

\subsection{Re-ID with Auxiliary Information}
Person image feature representations are affected by identity-unrelated factors, such as person pose, viewpoint, illumination, background, camera style, etc, which lead to large intra-class feature variance. Many researchers exploit auxiliary information to reinforce the feature representation. Example of auxiliary information include semantic attributes~\cite{ chen2021explainable}, viewpoint~\cite{liu2019view}, camera information~\cite{ zhu2019intra}. Some works use GAN to generate new body poses to strengthen robustness against pose variations~\cite{liu2018pose, zheng2019pose}. Though auxiliary feature learning methods have significantly boosted performance, these auxiliary information annotations are laborious and expensive except for camera labels. 

In practical surveillance systems, cameras are usually fixed and positioned in known locations, facilitating the acquisition of camera labels. Recent works have utilized camera labels to handle feature distribution discrepancy caused by the varied views and resolutions of different cameras, effectively improving performance. Some studies~\cite{zhong2018camera,lin2020crosscamera} adopt CyleGAN~\cite{zhu2017unpaired} to generate different camera styles images to mitigate the camera domain gap. Li~\cite{li2019unsupervised} exploits explicit cross-camera tracklet association to improve person tracking. CBN~\cite{zhuang2020rethinking} proposes the camera-based batch normalization to standardize feature vectors under different cameras, eliminating domain gaps under different cameras. SSL~\cite{lin2020unsupervised} introduces the cross-camera encouragement term to increase the disparity of image pairs from the same camera and minimize intra-camera negative pairs. CAP~\cite{wang2021camera} divides each cluster into multiple camera-aware proxies according to camera ID, capturing local structure within clusters to address intra-identity differences and inter-ID similarities. Based on CAP, O2CAP~\cite{wang2022offline} utilizes offline and online associations to reduce label noise and mine hard proxies. IICS~\cite{xuan2021intra} employs alternating intra-camera and inter-camera training to iteratively update the feature encoder. CaCL~\cite{lee2023cameradriven} establishes the camera-driven curriculum learning framework for progressively transferring knowledge from source to target domains. 

Among these approaches,~\cite{wang2021camera, wang2022offline, xuan2021intra, lee2023cameradriven} are most similar to ours. Their methods aim to utilize camera labels to optimize the global model across cameras. On the contrary, this work focuses on clustering intra-camera similarity and saving reliable local clustering results, which facilitates the refinement of global inter-camera pseudo labels. We utilize the refined pseudo labels for conducting effective learning in a self-paced way. 

\begin{figure*}[t!]
\centering
\includegraphics[width=15cm]{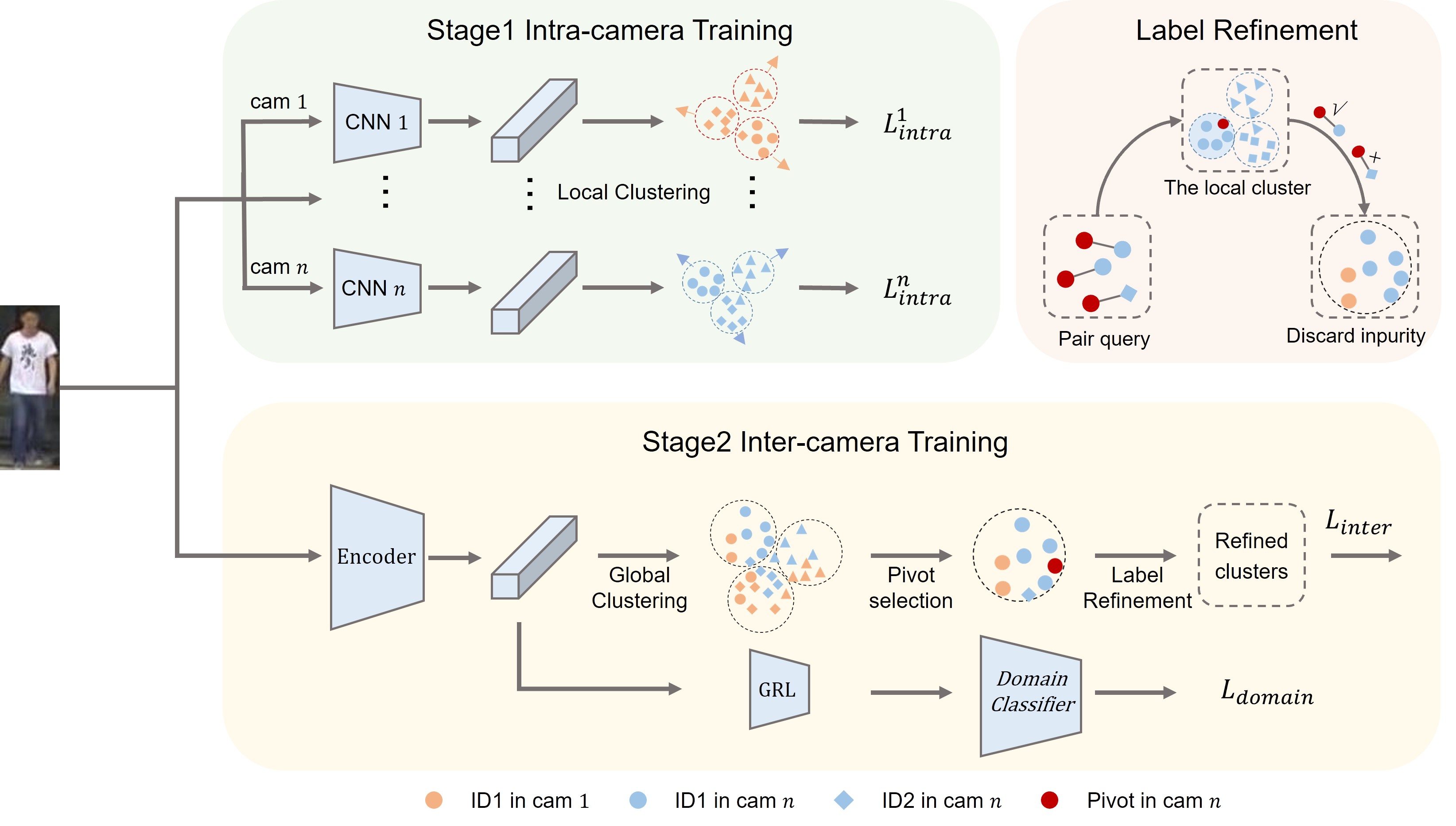}
\caption{The overview of our proposed CALR.  The intra-camera training stage optimizes each camera-specific CNN with local clustering and saves the final clustering results. The inter-camera training performs global clustering for all samples. Label refinement procedure exploits the reliable local cluster to estimate the pair relationship. The refined clusters are utilized to compute the inter-camera contrastive loss. We also perform camera domain classification on each feature embedding through the domain classifier and compute the domain classification loss.}
\label{img2}
\hspace{0.1in}
\end{figure*}

\section{Methodology} \label{sec3}
The proposed CALR addresses the cross-camera label noise and camera domain feature distribution discrepancy. Initially, we divide a target domain into multiple sub-domains by utilizing the camera labels for pedestrian images. Then we separately conduct intra-camera training for each sub-domain and obtain the reliable local clustering results. For the inter-camera training, we develop a novel criterion to select pivots to refine global clusters. Besides, we introduce a domain discriminator to align the camera domain feature distribution. Finally, we incorporate the inter-camera contrastive loss and domain classification loss to train the model progressively. Note that we only use camera information for training. During the inference time, we utilize pairwise distances between the query and gallery image features to retrieve the matched images under cross cameras. 

\subsection{Problem Formulation}
We denote the dataset without any identity annotations as $\mathcal{D}=\{x_i\}_{i=1}^N$, where $N$ denotes the total number of pedestrian images and $x_i$ denotes a pedestrian image. Assuming all the images are taken from $n$ disjoint cameras, and accordingly we divide $\mathcal{D}$ into $n$ non-overlapping sub-domains based on camera labels, $\mathcal{D}=\{\mathcal{D}^c\}_{c=1}^n$, where $\mathcal{D}^c$ denotes the sub-domain of person images under the same camera and $c=1 ... n$ is the index of cameras. The objective of person Re-ID task is to learn a discriminative and robust CNN encoder $f$ with parameters $\theta$ on $\mathcal{D}$. Given a query image $q$, the model $f$ is employed to extract features to retrieve the matching images from the gallery $\mathcal{G}$ under inter-cameras. We follow the most common pipeline in the clustering-based methods to learn a Re-ID model by alternating between the clustering and the model updating step. The parameters $\theta$ are initialized with the Imagenet~\cite{deng2009imagenet} pre-trained network. Since the number of person IDs is uncertain, we adopt Infomap~\cite{rosvall2008maps} to cluster the image features and assign images with ID labels. In this way, we get a labeled dataset $\mathcal{D}=\{x_i, y_i\}_{i=1}^N$, where $y_i\in\{0,1,2...,N_y\}$ is the generated pseudo label associated with image $x_i$ and $N_y$ denotes the total number of ID labels. With the ID labels, we could employ contrastive loss for model optimization. In this paper, we adopt the cluster-level contrastive loss~\cite{dai2022cluster}, which is defined by 
\begin{equation}
    \mathcal{L}_{base}=-log\frac{exp(q\cdot{u_+}/\tau)}{\sum_{k=0}^{K}exp(q\cdot{u_k}/\tau)}
    \label{e1}
\end{equation}
where $q$ is the query feature and $u_k$ is the cluster centroid defined by the mean feature vectors of each cluster. $u_+$ shares the same pseudo label with the query. $\tau$ is a temperature hyper-parameter. All cluster feature representation can be stored in a memory dictionary, which is updated consistently by corresponding query $q$ as:
\begin{equation}
    u_k = mu_k+(1-m)q
    \label{e2}
\end{equation}
where $m$ is the momentum updating factor. According to the baseline above, we optimize $f$ by two stages of training, consisting of the intra-camera training stage which clusters the intra-camera features and trains $C$ encoders $\{f^c\}_{c=1}^n$ respectively for each camera sub-domain, and the inter-camera training stage which refines pseudo labels with intra-camera clusters and trains model $f$. The overview of our framework for unsupervised person Re-ID is illustrated in Fig.~\ref{img2}, and the details will be discussed in the rest of the section.

\subsection{Intra-camera training}
As demonstrated in Fig.~\ref{img2}, the intra-camera training stage divides the training dataset into $n$ sub-sets. Because intra-camera feature distributions could escape from the influence of the camera domain gap, they are more concerned about the similarity of person appearances. With the intra-camera training in each sub-domain, we can obtain reliable local pseudo-labels for the next inter-camera training. Specifically, we adopt the pre-trained model to extract features for input images and perform clustering in each camera sub-domain $\mathcal{D}^c$ respectively. Images within the same cluster are assigned identical labels. With the generated pseudo labels, we get several labeled sub-sets $\{\mathcal{D}^c=\{x_i^c, y_i^c\}_{i=1}^{N^c}\}_{c=1}^n$, where $N^c$ is the total number of training samples of camera $c$.  We adopt the intra-camera contrastive loss for model updating. Given an image  $x_i^c$ captured from camera $c$, we use encoder $f^c$ to extract image feature $f^c(x^c_i)$ and compare it with all local cluster features stored in cluster-level memory $\mathcal{M}$ to compute loss value, which is express as 
\begin{equation}
    \mathcal{L}_{intra}^{c}=-log\frac{exp(f^c(x^c_i)\cdot{\mathcal{M}}(y_i^c)/\tau)} {\sum_{j=0}^{K^c}exp(f^c(x^c_i)\cdot{\mathcal{M}}(j)/\tau)}
    \label{e3}
\end{equation}
where $K^c$ denotes the total number of intra-camera clusters. The intra-camera contrastive loss pulls all instance features close to the corresponding cluster features and pushes them away from the other cluster features, which helps learn a specific feature encoder $f^c$ for each camera sub-domain. Note that the camera-specific encoders $\{f^c\}_{c=1}^n$ don't share any weights. Generally, image features depend on the person's appearance and other external factors. But images in the same camera sub-domain share the same settings of cameras including their parameters, viewpoint, resolution, environment, etc. Hence, the intra-camera features are more related to the characteristics of the person's identity. Based on that, the model training can have less performance degradation from label noise. The intra-camera step can reduce intra-identity variance and provide more reliable local pseudo labels. Stage 1 saves the final clustering results for the latter label refinement.

\subsection{Inter-camera training}
In the inter-camera training stage, we follow the intra-camera self-training scheme to cluster global features and assign identity labels to images across cameras. While the pre-trained feature encoder can learn general feature representations, the inter-camera feature distributions are heavily biased towards camera labels. Consequently, positive pairs obtained from different cameras may exhibit greater dissimilarity than negative samples from the same camera. As a result, identifying image pairs of the same identity across cameras and obtaining reliable pseudo-labels at the beginning of training stage becomes challenging, which leads to inevitable label noise. Therefore, the model is expected to initially learn from simple and reliable samples and gradually incorporate harder samples in a self-paced manner.

\begin{figure}[t!]
\centering
\includegraphics[width = \linewidth]{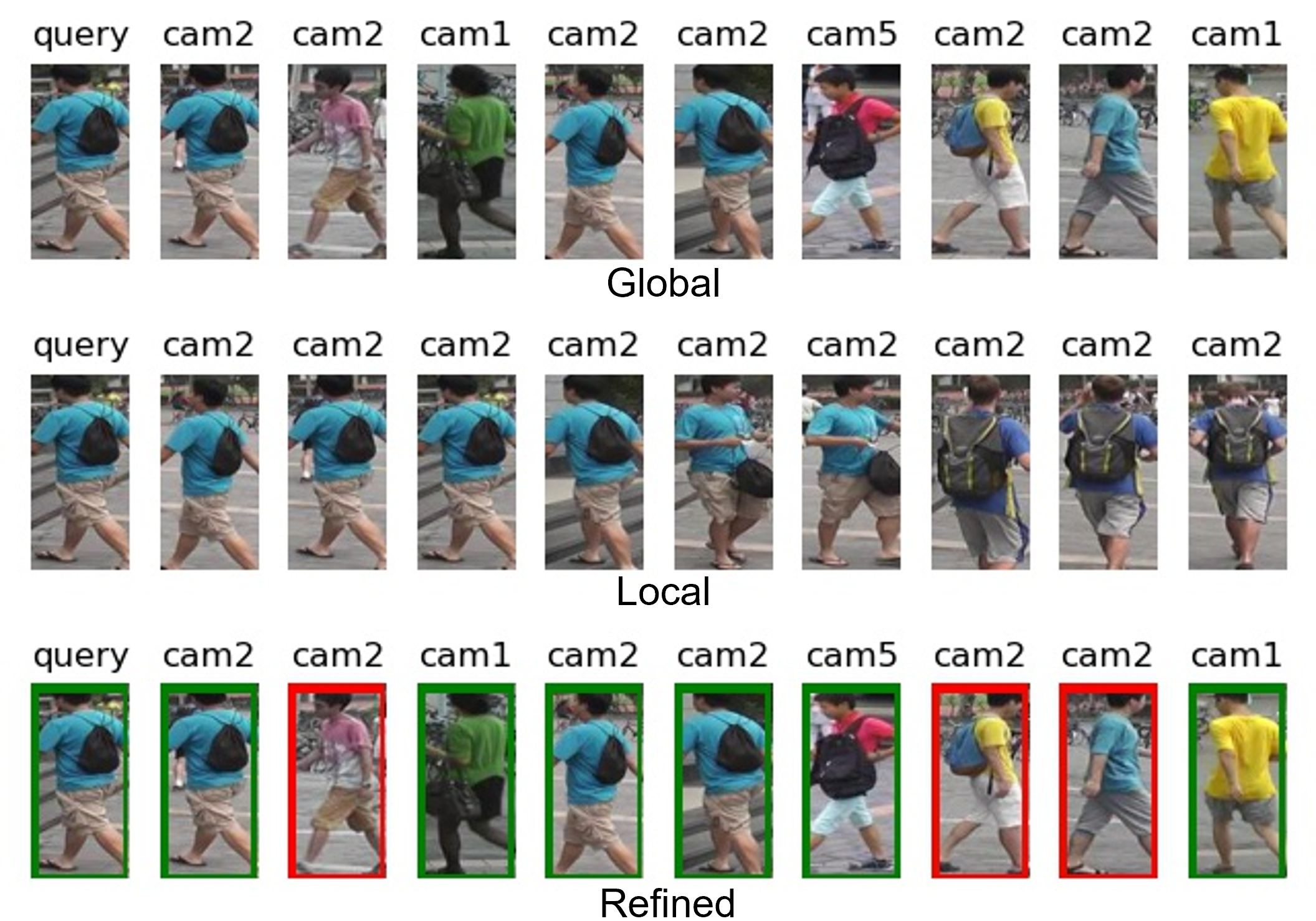}
\caption{Visualization of global cluster, local cluster, and refined cluster. Given a pivot, the first row denotes its global cluster and the second row denotes its local cluster. For the global cluster sample under the same camera with the pivot, we discard some samples which aren't clustered into the local cluster. The refined cluster is illustrated in the third row. Samples with red boxes are discarded, while those with green boxes do not.}
\label{img3}
\end{figure}

To address the problem above, we exploit reliable and fine-grained local clustering results as prior knowledge to refine the global pseudo labels. Due to the inherent limitations of intra-camera clustering, we can only assess image pair relationships within the same camera. At the onset of the training stage, intra-identity feature variance is quite large, leading to the presence of many nodes within each global cluster sharing the same camera ID, some of which may be inaccurately clustered. Our label refinement process aims to eliminate these impurities that can potentially degrade model performance. If we can identify pivot nodes with high utility within each cluster, the refinement process becomes straightforward. We only need to utilize the local clustering results to compare the pivot with other clustered nodes and discard negative samples. In essence, pivots are expected to represent the center of the cluster and be closely related to other clustered nodes. To identify these pivots, we design a criterion based on Harmonic Centrality~\cite{marchiori2000harmony} as follows: 
\begin{equation}
    score(i)=\sum_{j\in{{top}_{15}(i)}}\frac{1}{dist(i,j)+mean(dist)}
    \label{e4}
\end{equation}
where $dist(,)$ represents the distance between node $i$ and node $j$, and $mean(dist)$ denotes the mean distance across all pairs. The ${top}_{15}(i)$ refers to the top 15 closest neighbours of node $i$. A higher value of $score(i)$ indicates greater utility of node $i$. Therefore, we select samples with $score$ higher than the mean of all $score$ values as the pivots, which could be dynamically updated before each epoch.

Based on the selected pivots, we could refine the global pseudo labels. Given a pivot in camera $c$, locate the corresponding local cluster $L_i$ and the global cluster $G_k$. The global cluster $G_k$ can be subdivided into multiple sub-clusters $\{G_k^c\}_{c=0}^n$. For each node in $G_k^c$, we query the relationship between the pivot and other samples based on local clustering results. We retain the positive samples and discard the negative samples with a certain probability $p$. When the probability $p$ is set to $1$, the refined global cluster is defined as: 
\begin{equation}
    G_{refine}={G_k^c}\cap{L_i}\cup{(G_k\setminus G_k^c)}
    \label{e5}
\end{equation} where $G_k\setminus G_k^c$ is the set of all the global clustered samples except samples captured from camera $c$. Fig.~\ref{img3} is an example of label refinement at an early training epoch, where the query is a given pivot. As we can observe, label refinement effectively discards the impurities captured from the same camera based on local clustering result and improve the reliability of global clustering results. We utilize the refined pseudo labels to train the Re-ID backbone $f$. As discussed earlier, the model is anticipated to learn from initially reliable and simple samples to progressively harder samples. As the training continues, we can increase the difficulty of samples with refined labels by decaying the probability, which allows us to implement self-paced learning, gradually exposing the model to harder samples. can train the model with self-paced learning. After the label refinement, we obtain a training dataset $\mathcal{D}=\{x_i, y_i\}_{i=1}^N$ with reliable pseudo labels. Hence, the inter-camera contrastive loss is defined by
\begin{equation}
    \mathcal{L}_{inter}=-w_i\cdot{log\frac{exp(f(x_i)\cdot{\mathcal{M}}(y_i)/\tau)}{\sum_{j=0}^{K}exp(f(x_i)\cdot{\mathcal{M}}(j)/\tau)}}
    \label{e6}
\end{equation}where $K$ denotes the total number of refined inter-camera clusters. $w_i$ is the cluster-wise weighting factor. We calculate $w_i$ follows \cite{lee2023cameradriven}. The inter-camera contrastive loss enhances the person ID discrimination of the feature encoder within different cameras. 

\begin{algorithm}[t]
\caption{The overall training}\label{alg:alg1}
\begin{algorithmic}
\STATE {\textbf{Input: }}An unlabeled dataset $\mathcal{D}$, a initialized model $f$
\STATE {\textbf{Output: }} A trained Re-ID model
\STATE Stage 1: intra-camera training
\STATE \textbf{for} $c=1,...,n$ \textbf{do}
\STATE \hspace{0.3cm}Initialize $f^c$ with $f$
\STATE \hspace{0.3cm}\textbf{for} $epoch = 1,...,nums\_epochs$ \textbf{do}
\STATE \hspace{0.6cm}Extract features for $\mathcal{D}^c$ with $f^c$
\STATE \hspace{0.6cm}Clustering features and assigning pseudo labels
\STATE \hspace{0.6cm}Updating $f^c$ using Eq.\eqref{e3}
\STATE \hspace{0.3cm}\textbf{end}
\STATE \hspace{0.3cm}Save the final clustering results $L$
\STATE \textbf{end}
\STATE Stage 2: Inter-camera training
\STATE \textbf{for} $epoch = 1,...,nums\_epochs$ \textbf{do}
\STATE \hspace{0.3cm}Extract features for $\mathcal{D}$ with $f$
\STATE \hspace{0.3cm}Clustering features into global clusters $G$
\STATE \hspace{0.3cm}Select pivots with Eq.\eqref{e4}
\STATE \hspace{0.3cm}\textbf{for} i in pivots \textbf{do}
\STATE \hspace{0.6cm}Find global cluster $G_{i}$ and local cluster $L_i$
\STATE \hspace{0.6cm}Refine pseudo labels with Eq.\eqref{e5}
\STATE \hspace{0.3cm}\textbf{end}
\STATE \hspace{0.3cm}Updating $f$ using Eq.\eqref{e9}
\STATE \textbf{end}
\end{algorithmic}
\label{alg}
\end{algorithm}

In essence, our label refinement is to select more reliable samples for model updating to alleviate label noise. However, it doesn't explicitly align the distribution of the camera sub-domain. To overcome the feature distribution discrepancy, we propose a camera domain alignment module to pull different camera domains together. In the inter-camera training stage, we introduce an auxiliary task to perform domain classification on each feature embedding through the domain classifier and determine which camera domains it comes from. On the trained domain classifier, it is assumed that the features from different domains cannot correctly distinguish which one comes from the domain, in other words, the classification loss is large, and then the encoding feature is the camera-invariant feature. Therefore, the domain classifier and the feature encoder form adversarial training. Continuously minimizing inter-camera contrastive loss of the main task and maximizing the camera domain classification loss of the auxiliary task, the final learned feature representation is both discriminative and camera-invariant. To make the network conform to the standard forward propagation, we utilize a special Gradient Reverse Layer (GRL)~\cite{ganin2016domain} inserted between the feature encoder and the camera domain classifier. GRL has two unequal representations in the forward and backward propagation, which is defined by
\begin{equation}
    \mathcal{R}_\lambda(x) = x, \quad   
    \frac{d\mathcal{R}_\lambda(x)}{dx} = -\lambda I
    \label{e7}
\end{equation} where $\lambda$ is the only parameter. During forward propagation, GRL functions as an identity transform. During backward propagation, the gradient reverse and multiple $\lambda$. Based on the GRL, we could compute the camera domain classification loss as follows
\begin{equation}
    \mathcal{L}_{domain} = -\sum_{i=0}^{N}{c_i}\cdot log(\mathcal{D}(\mathcal{R}_\lambda(f(x_i)))
    \label{e8}
\end{equation}where$x_i$ is a pedestrian image and  $c_i$ is the matching camera label. $\mathcal{D}$ is the domain classifier. We train the domain classifier with adversarial learning to align the camera domain feature distributions.

\subsection{Overall training}
The overall Re-ID model is optimized using two stages of training. Stage 1 obtains reliable and fine-grained local pseudo labels for each camera sub-domain. Based on the results of Stage 1, Stage 2 conducts label refinement to mitigate the label noise stemming from the camera bias. Besides, we introduce camera domain classification as an auxiliary task to alleviate the feature distribution discrepancy. We jointly adopt inter-camera contrastive loss and camera domain classification loss for the inter-camera training. In summary, the overall objective can be formulated as follows
\begin{equation}
    \mathcal{L} = \mathcal{L}_{inter} + \beta \cdot \mathcal{L}_{domain}
    \label{e9}
\end{equation}where $\beta$ is the hyper-parameter used to balance the two losses. Algorithm~\ref{alg} is an  an outline of the overall training process of our approach.

\begin{table*}[t!]
\caption{Comparison of State-of-The-Arts Methods on person datatsets. The \textbf{best} result is denoted in bold black, while the \underline{second-best} result is underlined. \label{table1}}
\centering
\renewcommand{\arraystretch}{1.8}
\resizebox{\linewidth}{!}{
\begin{tabular}{cc|cccc|cccc|cccc}
\hline
\multirow{2}{*}{Methods}& \multirow{2}{*}{Reference}&
\multicolumn{4}{c|}{Market1501}& \multicolumn{4}{c|}{DukeMTMC-ReID}& 
\multicolumn{4}{c}{MSMT17}\\
\cline{3-14}
&&mAP&rank-1&rank-5&rank-10&mAP&rank-1&rank-5&rank-10&mAP&rank-1&rank-5&rank-10 \\
\hline
\multicolumn{14}{l}{\textbf{Supervised}}\\
\hline
% BOT\cite{luo2019bag}&CVPRW19&94.2&95.4&-&-&89.1&90.3&-&-&-&-&-&-\\
TransReID\cite{he2021transreid}&ICCV21&89.5&96.2&-&-&82.6&90.7&-&-&69.4&86.2&-&-\\
%Nformer\cite{wang2022nformer}&CVPR22&93.0&95.7&-&-&85.7&90.6&-&-&62.2&80.8&-&-\\
Fastreid\cite{he2023fastreid}&ACMMM23&90.3&96.3&-&-&83.2&92.4&-&-&63.3&63.3&-&-\\
\hline
\multicolumn{14}{l}{\textbf{Unsupervised domain adaptation}}\\
\hline
% MMT(IBN)\cite{ge2020mutual}&ICLR20&76.5&90.9&96.4&97.9&68.7&81.8&91.2&93.4&29.7&58.8&71.0&76.1\\
%SPCL\cite{ge2020self}&NeurIPS20&77.5&89.7&96.1&97.6&68.8&82.9&90.1&92.5&26.8&53.7&65.0&69.8\\
SPCL(IBN)\cite{ge2020self}&NeurIPS20&79.2&91.5&96.9&98.0&69.9&83.4&91.0&93.1&31.8&58.9&70.4&75.2\\
% DARC\cite{hu2022divide}&AAAI22&85.1&94.1&97.6&98.7&-&-&-&-&35.2&64.5&76.2&80.4\\
% AWB\cite{wang2022attentive}&TIP22&81.0&93.5&97.4&98.3&70.9&83.8&92.3&94.0&29.5&61.0&73.5&77.9\\
CACHE\cite{liu2022complementary}&TCSVT22&83.1&93.4&97.5&98.2&\underline{71.7}&83.5&91.4&93.9&31.3&58.0&69.8&74.5\\
% Fastreid(GemPool)\cite{he2023fastreid}&ACMMM23&80.5&92.7&-&-&69.2&82.7&-&-&27.7&59.5&-&-\\
% CaCL\cite{lee2023cameradriven}&ICCV23&84.7&93.8&97.7&98.6&-&-&-&-&36.5&66.6&75.3&80.1\\
FUReID\cite{peng2024adapt}&PR24&81.7&92.7&97.2&98.9&-&-&-&-&26.2&53.6&64.1&68.6\\
IDENet\cite{yang2025idenet}&TIP25&83.6&93.3&97.6&98.4&-&-&-&-&31.9&60.6&71.5&75.9\\
\hline
\multicolumn{14}{l}{\textbf{Purely unsupervised without camera labels}}\\
\hline
BUC\cite{lin2019bottom}&AAAI19&38.3&66.2&79.6&84.5&27.5&47.4&62.6&68.4&-&-&-&-\\
SSL\cite{lin2020unsupervised}&CVPR20&37.8&71.7&83.8&87.4&28.6&52.5&63.5&68.9&-&-&-&-\\ 
% MMCL\cite{wang2020unsupervised}&CVPR20&45.5&80.3&89.4&92.3&40.2&65.2&75.9&80.0&11.2&35.4&44.8&49.8\\
RLCC\cite{zhang2021refining}&CVPR21&77.7&90.8&96.3&97.5&69.2&83.2&91.6&93.8&27.9&56.5&68.4&73.1\\
% ICE\cite{chen2021ice}&CVPR21&79.5&92.0&97.0&98.1&67.2&81.3&90.1&93.0&29.8&59.0&71.7&77.0\\

% HCM\cite{si2022hybrid}&TMM22&79.0&91.8&96.7&97.7&67.9&82.3&90.2&92.8&26.9&59.6&70.1&74.3\\
PPLR\cite{cho2022part}&CVPR22&81.5&92.8&97.1&98.1&-&-&-&-&31.4&61.1&73.4&77.8\\
% ISE(GeMPool)\cite{zhang2022implicit}&CVPR22&85.3&94.3&98.0&98.8&-&-&-&-&37.0&67.6&77.5&81.0\\
ClusterComtrast\cite{dai2022cluster}&ACCV22&83.0&92.9&97.2&98.0&-&-&-&-&33.0&62.0&71.8&76.7\\
RPE\cite{wang2023relation}&TMM23&82.4&92.6&97.1&97.9&71.5&77.8&89.3&91.7&-&-&-&-\\
% DiDAL\cite{liu2023discriminative}&TMM23&84.8&94.2&98.2&-&-&-&-&-&45.4&74.0&84.3&-\\
HCACE\cite{luo2024hierarchical}&TMM24&83.4&\textbf{93.7}&\underline{}&98.1&71.5&\underline{84.2}&\underline{91.9}&\textbf{94.2}&41.6&72.4&81.8&84.9\\
\hline
\multicolumn{14}{l}{\textbf{Purely unsupervised with camera labels}}\\
\hline
CAP\cite{wang2021camera}&AAAI21&79.2&91.4&96.3&97.7&67.3&81.1&89.3&91.8&36.9&67.4&78.0&81.4\\
IICS\cite{xuan2021intra}&CVPR21&72.9&89.5&95.2&97.0&64.4&80.0&89.0&91.6&26.9&56.4&68.8&73.4\\
% ICE\cite{chen2021ice}&CVPR21&82.3&93.8&97.6&98.4&69.9&83.3&91.5&94.1&38.9&70.2&80.5&84.4\\
IIDS\cite{xuan2022intra}&TPAMI22&78.0&91.2&96.2&97.7&68.7&82.1&90.8&93.7&35.1&64.4&76.2&80.5\\
PPLR\cite{cho2022part}&CVPR22&84.4&92.8&97.1&98.1&-&-&-&-&42.2&\underline{73.3}&\underline{83.5}&\underline{86.5}\\
%O2CAP\cite{wang2022offline}&TIP22&82.7&92.5&96.9&98.0&71.2&83.9&91.3&93.4&42.4&72.0&81.9&85.4\\
%O2CAP(IBN)\cite{wang2022offline}&TIP22&83.7&93.1&97.4&98.1&72.8&85.2&91.9&93.5&46.9&75.5&84.8&87.7\\
% CIFL\cite{pang2022camera}&TMM22&82.4&93.9&97.9&98.1&69.8&83.7&91.8&94.2&38.8&70.1&80.7&83.9\\
% O2CAP(IBN+GeMPool)\cite{wang2022offline}&TIP22&85.0&93.3&96.9&97.7&73.2&85.3&91.4&93.3&48.3&77.3&85.6&88.2\\
CC+CAJ\cite{yiyu2024caj}&CVPR24&\textbf{84.8}& \underline{93.6}& \textbf{97.6}& \textbf{98.4}&-&-&-&-& \underline{42.8}& 72.3& 82.2& 85.6\\
% \rowcolor{lightgray}
% Baseline&This work&78.7&90.7&96.3&97.3&68.6&81.6&90.7&93.2&33.6&62.4&72.7&77.1\\
% \rowcolor{lightgray}
% CALR&This work&82.8&92.5&96.9&98.0&71.5&82.7&91.6&93.9&44.2&73.0&82.6&85.9\\
% \rowcolor{lightgray}
% CALR(IBN)&This work&83.7&93.3&96.7&97.8&72.5&85.2&91.8&93.8&47.9&75.9&85.4&88.4\\
% \rowcolor{lightgray}
% CALR(IBN+GeMPool)&This work &84.5&93.6&97.5&98.3&74.2&86.0&92.3&94.2&50.4&78.1&86.4&89.4\\
CALR(Ours)&This work &\underline{84.5}&\underline{93.6}&\underline{97.5}&\underline{98.3}&\textbf{74.2}&\textbf{86.0}&\textbf{92.3}&\textbf{94.2}&\textbf{50.4}&\textbf{78.1}&\textbf{86.4}&\textbf{89.4}\\
\hline
\end{tabular}
}
\end{table*}

\section{Experiments Results} \label{sec4}

\subsection{Implementation Details}
{\bf{Datasets and Evaluation Protocols. }}Our proposed CALR was assessed using three widely used person re-identification datasets: Market1501~\cite{zheng2015scalable}, DukeMTMC-reID~\cite{ristani2016performance}, and MSMT17~\cite{wei2018person} respectively.
These three datasets are collected from real-world surveillance scenarios. Market1501 comprises
32668 images of 1501 pedestrian identities collected by 6 disjoint cameras. DukeMTMC-reID
has 16,522 images of 702 pedestrian identities obtained from 8 different cameras. MSMT17
contains 126441 images of 4101 pedestrian identities capture from 15 cameras. It covers different
time in 4 days with different weather. To validate the generalization capacity of our proposed
CALR, we further evaluated it on a vehicle Re-ID dataset, Veri-776~\cite{liu2017provid}, comprising over 50,000
images of 776 vehicles collected by 20 non-overlapping cameras. It is captured from the real
traffic scenario. 

For evaluation, we utilized two widely adopted metrics: Cumulative Matching Characteristic (CMC) at Rank-k and mean average precision (mAP).

{\bf{Training details. }} We build our CALR on the baseline ClusterContrast~\cite{dai2022cluster}. Specifically, we adopt ResNet-50~\cite{he2016deep}, pre-trained on ImageNet~\cite{deng2009imagenet} classification, as the backbone for extracting feature. Particularly, we used the generalized mean pooling layer (GemPool) followed by the instance batch normalization (IBN)~\cite{ulyanov2016instance} layer and L2-normalization layer instead of the fully connected classification layer to output 2048-dimensional features. The input image was resized to 256 x 128 for person Re-ID datasets and 224 × 224 for Veri-776. The memory updating factor $m$ was 0.2. The temperature hyper-parameter $\mathcal{\tau}$ was set to $0.1$. The model was trained for 20 epochs for the intra-camera training and 50 epochs for the inter-camera training. We performed the Agglomerative Hierarchical~\cite{pedregosa2011scikit} clustering for the intra-camera step, with the number of clusters being $num{\_}images/5$ for each camera. In the inter-camera step, we used the two-stage Infomap method~\cite{rosvall2008maps} for clustering. The training optimizer was Adam with 5e-4 weight decay.

% {\bf{Comparison with more advanced backbones. }}Recent works~\cite{ge2020mutual, wang2022offline, he2023fastreid} utilize instance batch normalization (IBN)~\cite{ulyanov2016instance} replace BN layer to enhance the model generalization. The experimental results presented in Table~\ref{table1} and Table~\ref{table2} show the effectiveness of our proposed method combined with IBN-ResNet50, which validates the significant boost of IBN in enhancing the Re-ID model performance. Using the generalized mean pooling layer (GeMPool) instead of GAP layer also provides a significant performance increase. CALR with IBN-ResNet50 and GeMPool surpasses CALR on all the benchmarks with +1.7$\%$, +2.7$\%$, +6.2$\%$ and +1.9$\%$ in mAP on Market1501, DukeMTMC-ReID, MSMT17 and Veri-776, respectively. It proves that our proposed CALR can adapt well to the more advanced backbones.

\begin{table}[t]
\caption{Comparison of State-of-The-Arts Methods on Veri-776.\label{table2}}
\centering
\renewcommand{\arraystretch}{2}
\resizebox{\linewidth}{!}{
\begin{tabular}{cc|cccc}
\hline
\multirow{2}{*}{Methods}& \multirow{2}{*}{Reference}&
\multicolumn{4}{c}{Veri776}\\
\cline{3-6}
&&mAP&rank-1&rank-5&rank-10 \\
\hline
\multicolumn{6}{l}{\textbf{Unsupervised domain adaptation}}\\
\hline
MMT\cite{ge2020mutual}&ICLR20&35.3&74.6&82.6&87.0\\
SPCL\cite{ge2020self}&NeurIPS20&38.9&80.4&86.8&89.6\\
% AWB\cite{wang2022attentive}&TIP22&37.2 &79.9 &85.2 &89.2\\
\hline
\multicolumn{6}{l}{\textbf{Purely unsupervised without camera labels}}\\
\hline
SPCL\cite{ge2020self}&NeurIPS20&36.9 &79.9 &86.8 &89.9\\
RLCC\cite{zhang2021refining}&CVPR21&39.6&83.4 &88.8 &90.9\\
PPLR\cite{cho2022part}&CVPR22&41.6&85.6&91.1&93.4\\
ClusterContrast\cite{dai2022cluster}&ACCV22&40.8&86.2&90.5&92.8\\
%O2CAP\cite{wang2022offline}&TIP22&41.9&87.5&92.7&94.4\\
%O2CAP(IBN)\cite{wang2022offline}&TIP22&42.4&89.6&93.5&94.7\\
% O2CAP(IBN+GeMPool)\cite{wang2022offline}&TIP22&43.0&89.7&93.8&95.1\\
% DiDAL\cite{liu2023discriminative}&TMM23&43.5&89.0&93.5&-\\
\hline
\multicolumn{6}{l}{\textbf{Purely unsupervised with camera labels}}\\
\hline
PPLR\cite{cho2022part}&CVPR22&43.5 &88.3 &92.7 &94.4\\
CC+CAJ\cite{yiyu2024caj}&CVPR24&43.1& 90.1& 92.8 &\underline{95.0}\\
CDF\cite{qiu2024camera}&TCSVT24&\underline{44.0}&\underline{90.5}&\underline{93.7}&94.9\\
% \hline
% \rowcolor{lightgray}
% Baseline&This work&38.8&85.9&90.6&93.0\\
% \rowcolor{lightgray}
% CALR&This work&42.9&88.7&92.6&94.8\\
% \rowcolor{lightgray}
% CALR(IBN)&This work&43.9&91.2&93.6&94.9\\
% \rowcolor{lightgray}
CALR&This work&\textbf{44.8}&\textbf{91.6}&\textbf{93.9}&\textbf{95.2}\\
\hline
\end{tabular}
}
\end{table}

\subsection{Comparison with the State-of-the-Art Methods}
We conducted a comparative analysis of our proposed method against state-of-the-art works, encompassing supervised Re-ID, UDA Re-ID and purely unsupervised Re-ID with and without camera labels. Table~\ref{table1} summarizes the comparison results on three pedestrian datasets. Table~\ref{table2} shows the evaluation results on a vehicle dataset. 

{\bf{Comparison with supervised person Re-ID methods. }}Table ~\ref{table1} illustrates the advanced fully supervised Re-ID works including TransReID~\cite{he2021transreid}, and Fastreid~\cite{he2023fastreid}. Due to the lack of ID labels, there exists a notable performance gap between the unsupervised and fully supervised methods. Even if unsupervised methods get poor performance, our proposed CALR framework achieves considerable improvement to mitigate the gap, showing the scalability in real-world deployments. %Our proposed method aims to narrowing this gap.

{\bf{Comparison with UDA person Re-ID methods. }}We provide several recent unsupervised domain adaptation works for comparison, including SPCL~\cite{ge2020self}, CACHE~\cite{liu2022complementary}, FUReID~\cite{peng2024adapt} and IDENet~\cite{yang2025idenet}. Despite UDA-based methods utilizing external annotation to enhance Re-ID performance, they do not demonstrate significant improvement. Without any identity annotation, a purely unsupervised setting poses a significantly more challenging task. Despite this, our proposed method showcases the superior performance, outperforming UDA methods by a large margin, which indicates the capacity of CALR to effectively leverage unlabeled data and explore valuable information.%on the three datatsets, especially on MSMT17 with complex lighting and scene variations.

{\bf{Comparison with purely unsupervised person Re-ID methods. }}We divided the purely unsupervised methods into two categories based on whether camera labels were used or not. Most of these camera-agnostic methods, including 
BUC~\cite{lin2019bottom}, 
SSL~\cite{lin2020unsupervised}
RLCC~\cite{zhang2021refining},
PPLR~\cite{cho2022part}, 
% ISE~\cite{zhang2022implicit}, 
ClusterContrast~\cite{dai2022cluster}, RPE~\cite{wang2023relation} and HCACE~\cite{luo2024hierarchical} exploit robust clustering methods to generate accurate labels and design effective strategies to reduce label noise. Without camera information, it is difficult for them to cope with the label noise caused by camera domain shifts. They therefore demonstrate poor performance in large and challenging datasets. Compared with those works, unsupervised methods using camera labels~\cite{wang2021camera, xuan2021intra,xuan2022intra, cho2022part, yiyu2024caj} show more competitive performance. However, all these camera-aware methods aim to utilize camera labels to optimize the global Re-ID model under different cameras. Our CALR focuses on the reliable and fine-grained local labels in each camera and employs them to refine global labels across cameras, which effectively reduces the label noise. As shown in Table ~\ref{table1}, our CALR demonstrates promising result with \textbf{mAP = 84.5$\%$} and \textbf{rank-1 = 93.6$\%$} on Market1501, while our methods significantly surpasses prior stat-of-the-art methods with \textbf{mAP = 74.2$\%$} and \textbf{rank-1 = 86.0$\%$} on DukeMTMC-ReID; and \textbf{mAP = 50.4$\%$} and \textbf{rank-1 = 78.1$\%$} on MSMT17. 

% {\bf{Comparison with more advanced backbones. }}Recent works~\cite{ge2020mutual, wang2022offline, he2023fastreid} utilize instance batch normalization (IBN)~\cite{ulyanov2016instance} replace BN layer to enhance the model generalization. The experimental results presented in Table~\ref{table1} and Table~\ref{table2} show the effectiveness of our proposed method combined with IBN-ResNet50, which validates the significant boost of IBN in enhancing the Re-ID model performance. Using the generalized mean pooling layer (GeMPool) instead of GAP layer also provides a significant performance increase. CALR with IBN-ResNet50 and GeMPool surpasses CALR on all the benchmarks with +1.7$\%$, +2.7$\%$, +6.2$\%$ and +1.9$\%$ in mAP on Market1501, DukeMTMC-ReID, MSMT17 and Veri-776, respectively. It proves that our proposed CALR can adapt well to the more advanced backbones.

{\bf{Comparison with SOTA methods on vehicle Re-ID. }}The comparison results are dipicted in Table~\ref{table2}. The UDA-based methods including MMT~\cite{ge2020mutual}, SPCL~\cite{ge2020self}. The purely unsupervised methods including SPCL~\cite{ge2020self}, RLCC~\cite{zhang2021refining}, PPLR~\cite{cho2022part}, ClusterContrast~\cite{dai2022cluster}, O2CAP \cite{wang2022offline}, DiDAL~\cite{liu2023discriminative} and CAJ~\cite{yiyu2024caj}. Our proposed method achieves significant performance improvement with \textbf{mAP = 44.8$\%$} and \textbf{rank-1 = 91.6$\%$}, outperforming the baseline by a remarkable margin. Compared with the state-of-the-art methods, the proposed CALR maintains a competitive advantage, which validates its great availability in vehicle Re-ID.

\begin{figure}[t!]
    \centering
    \includegraphics[width=0.8\linewidth]{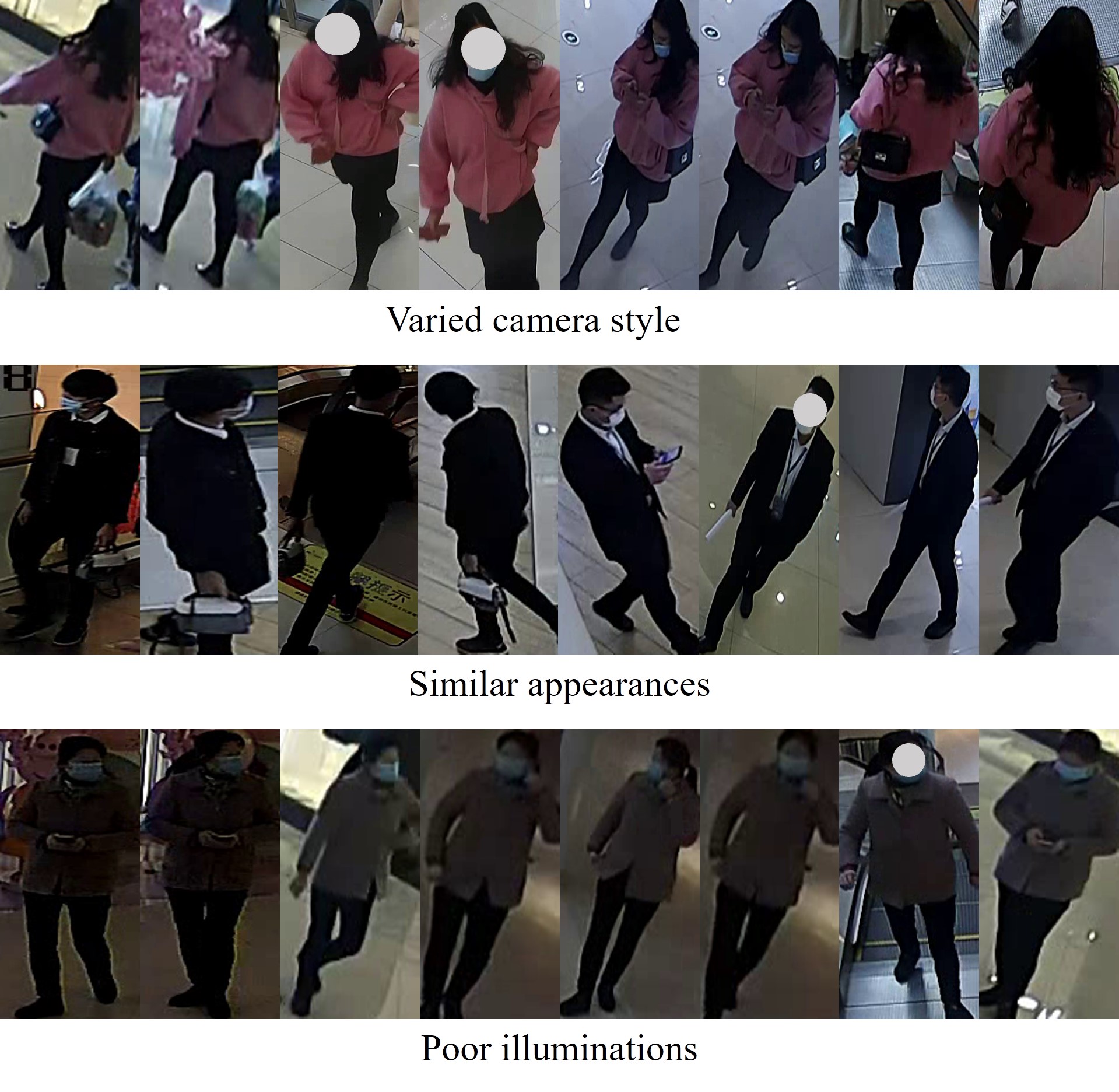}
    \caption{The examples of images and challenge of the self-collected real-world dataset.}
    \label{fig:nbhy}
\end{figure}

\begin{table}[t!]
\caption{Experiments on a self-collected real-world dataset.\label{table_NBHY}}
\centering
\renewcommand{\arraystretch}{1.6}
\resizebox{\linewidth}{!}{
\begin{tabular}{cc|cccc}
\hline
Methods&referemce&mAP&rank-1&rank-5&rank-10\\
\hline
SPCL~\cite{ge2020self}&Neurips20&24.5&30.4&39.7&51.4\\
ClusterContrast~\cite{dai2022cluster}&ACCV22&32.3&33.6&48.6&57.0\\
CC+CAJ~\cite{yiyu2024caj}&CVPR24&35.3&44.9&61.2&72.9\\
CALR&this paper&\textbf{45.7}&\textbf{50.5}&\textbf{65.4}&\textbf{73.4}\\
\hline
\vspace{-3mm}
\end{tabular}
}
\end{table}

\subsection{Experiments on self-collected real-world dataset}
To further explore the real-world applicability of our proposed method, we collected a new dataset from a shopping mall. This setting is particularly relevant for retail analytics and closer to real-world scenarios than typical laboratory settings. The new dataset is composed of 5465 pedestrian images of 555 pedestrian identities captured from 14 non-overlapping cameras. For evaluation, the dataset is divided into 3353 images of 255 identities for training, 289 query images and 1823 gallery images of 150 identities for testing. We provide some example images of this dataset in Figure \ref{fig:nbhy}, which shows it is a challenging dataset captured from a complex indoor environment. The challenges presented by this dataset are manifold: 
\textit{(a) Varied camera style}: Pedestrian images are captured from different cameras with diverse viewpoints, poses and resolutions. It's difficult to maintain consistent identification under the camera variation. 
\textit{(b) Similar appearances among pedestrians}: A notable difficulty in this environment is the similarity in appearance among pedestrians, particularly due to the uniforms worn by staff. This similarity poses a unique challenge in distinguishing between different individuals.
\textit{(c) Poor illumination}: Being an indoor dataset, the variability in lighting conditions further complicates the task of person re-identification, affecting the quality of the captured images. Due to the various challenges mentioned, it's tough to learn consistent representation for the same person with an unsupervised learning strategy.

To better validate the effectiveness of our method in practical application, we compare our method with popular SOTA methods SPCL~\cite{ge2020self}, ClusterContrast~\cite{dai2022cluster} and CAJ~\cite{yiyu2024caj} on the self-collected dataset. Table~\ref{table_NBHY} illustrate the comparison results. Our method demonstrates superior performance. This experiment validates the great availability of our approach in the real-world scenario.

\begin{figure}[t!]
\centering
\includegraphics[width = \linewidth]{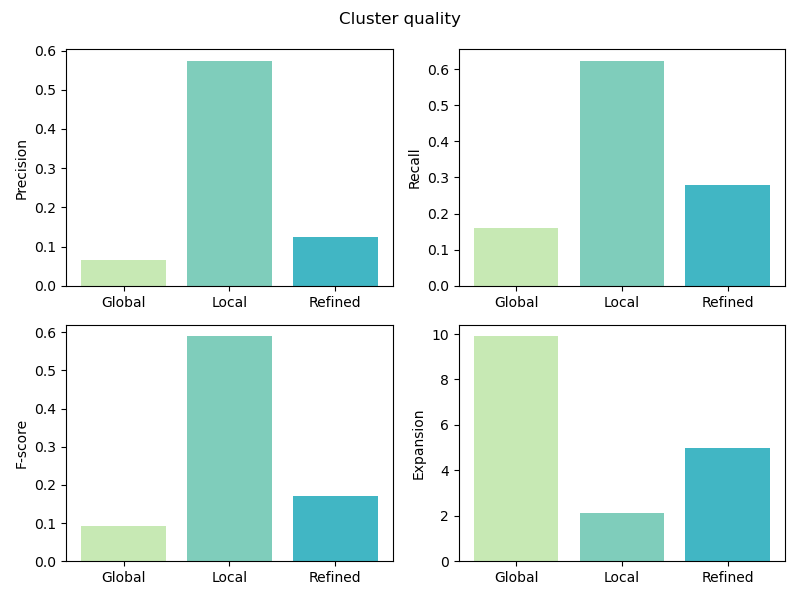}
\caption{Comparison of clustering quality in the global clusters, the local clusters, and the refined clusters. We utilize precision, recall, f-score, and expansion metrics to analyze the clusters. Expansion refers to the average number of clusters to which an ID is classified. The global clusters are obtained from inter-camera clustering on Market1501. The local clusters are the final clustering results of the intra-camera training stage.}
\label{img4}
\end{figure}

\begin{table*}[!t]
\caption{Ablation study on individual components of our proposed method.\label{table3}}
\centering
\renewcommand{\arraystretch}{1.6}
\resizebox{\linewidth}{!}{
\begin{tabular}{|ccc|cccc|cccc|cccc|}
\hline
\multirow{2}{*}{Model}& 
\multicolumn{2}{c|}{Component}&
\multicolumn{4}{c|}{Market1501}& \multicolumn{4}{c|}{DukeMTMC-ReID}& 
\multicolumn{4}{c|}{MSMT17}\\
\cline{2-15}
&$CA$&$LR$&mAP&rank-1&rank-5&rank-10&mAP&rank-1&rank-5&rank-10&mAP&rank-1&rank-5&rank-10 \\
\hline
Baseline&&& 83.0&92.9&97.2&98.0& 72.6&84.7&91.0&93.2&40.0&70.0&79.7&83.0 \\
Ours&\checkmark&&83.9&93.0&97.1&98.2& 73.1&84.9&91.5&93.6&42.1&72.2&81.7&84.8 \\
%Ours&&\checkmark&&\textbf{84.4}&93.2&97.2&98.2& 73.8&85.6&92.2&94.3& 46.6&74.8&83.7&86.7 \\
%Ours&&&\checkmark&84.3&93.0&97.2&98.0 &73.5&85.1&92.1&94.1 &48.2&76.6&85.7&88.3\\
%Ours&\checkmark&\checkmark&& 84.3&\textbf{93.6}&\textbf{97.4}&\textbf{98.3}& 73.6&\textbf{85.9}&92.2&93.5&45.7&74.0&83.3&86.4\\
Ours&&\checkmark&84.4&93.2&97.2&98.2 &73.8&85.2&92.1&94.1 &49.6&77.5&86.1&88.8\\
%Ours&\checkmark&&\checkmark&84.2&93.3&97.3&98.1 &73.5&85.3&91.9&93.6 &&&&\\
Ours&\checkmark&\checkmark&\textbf{84.5}&\textbf{93.6}&\textbf{97.5}&\textbf{98.3}& \textbf{74.2}&\textbf{86.0}&\textbf{92.3}&\textbf{94.2}& \textbf{50.4}& \textbf{78.1}& \textbf{86.4}& \textbf{89.4} \\
\hline
\end{tabular}
}
\end{table*}

\subsection{Ablation studies}
The following subsections systematically investigate the effectiveness of our proposed camera-aware label refinement framework including label refinement with intra-camera clustering and camera domain alignment module. We first compared the quality of clusters to validate the effectiveness of our label refinement, and the result is reported in Fig.~\ref{img4}. Then, we conducted a thorough ablation analysis for each proposed component. We show mAP and rank-k scores to estimate the performance of baseline and our model with different components in Table~\ref{table3}, where $CA$ is the camera domain alignment module. $LR$ is the label refinement module. Moreover, we investigated the validation of the probability decaying strategy in the inter-camera training step.

\begin{figure}[t]
\centering
\includegraphics[width = 0.95\linewidth]{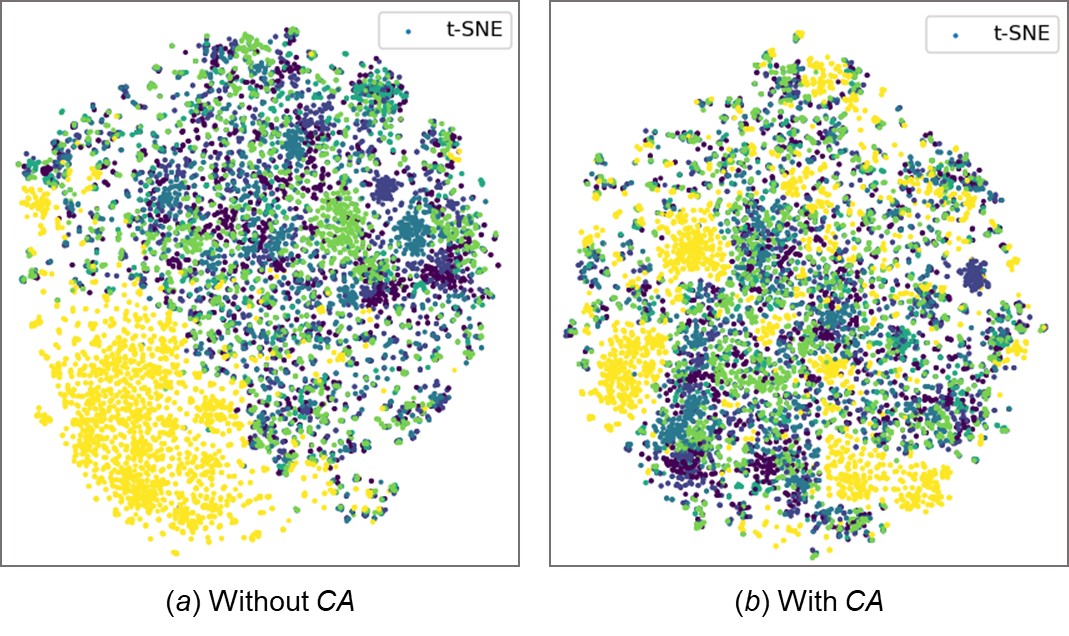}
\caption{T-SNE visualization of feature distribution of our method without and with camera domain alignment module, where features are extracted from Market-1501 at the beginning of training process. Different colors denotes samples from different cameras. The left figure shows the domain gap still exists in different camera domains. The right figure illustrates that different camera domains have more similar distributions. }
\label{img5}
\end{figure}

{\bf{Evaluation of clustering quality. }}To prove the necessity of the label refinement, we first evaluated the clustering quality of global, local, and refined clusters on Market1501 in Fig.~\ref{img4}. We adopt several common clustering evaluation metrics including precision, recall, f-score, and expansion to analyze the clustering quality. Expansion denotes the total number of clusters to which images of an ID are classified. Fig.~\ref{img4} shows that the quality of local clusters is far better than global clusters. After the label refinement, the expansion has been reduced to half its initial number and other metrics are significantly superior to the global clusters, which suggests label refinement discards some hard negative samples that are misclassified into other clusters. Our label refinement enhance the accuracy and reliability of the global clusters, validating the excellence of the label refinement framework in handling label noise.

{\bf{Effectiveness of the label refinement. }}Through the comprehensive experimental results in Table~\ref{table3}, we investigate the validation of the label refinement module in enhancing the model performance. These experiments were conducted in three benchmark datasets. We employ our proposed method with and without the label refinement module. From the comparison of Table~\ref{table3}, we observe our method with label refinement significantly outperforms the baseline in all datasets, particularly on the large and challenging MSMT17. It demonstrates the effectiveness and superiority of label refinement. 

\begin{table}[!t]
\caption{Ablation study on decay strategy of our proposed method.\label{table4}}
\centering
\renewcommand{\arraystretch}{1.5}
\resizebox{8cm}{!}{
\begin{tabular}{|c|cc|cc|}
\hline
\multirow{2}{*}{Decaying}&
\multicolumn{2}{c|}{Market1501}& \multicolumn{2}{c|}{DukeMTMC-ReID}\\
\cline{2-5}
&mAP&rank-1&mAP&rank-1\\
\hline
No Decay&83.6&92.8& 74.1&85.8\\
Linear&83.7&93.3 &73.7&85.9 \\
Polynomial&83.1&92.4&73.5&85.8\\
Exponential&83.9&93.5&73.4&85.2\\
Cosine&\textbf{84.5}&\textbf{93.6}&\textbf{74.2}&\textbf{86.0} \\
\hline
\end{tabular}
}
\end{table}

\begin{table}[!t]
\caption{The performance evaluation of hyper-parameter $\beta$.\label{table5}}
\centering
\renewcommand{\arraystretch}{1.25}
\resizebox{7.5cm}{!}{
\begin{tabular}{c|cc|cc}
\hline
\multirow{2}{*}{$\beta$}&
\multicolumn{2}{c|}{Market1501}&\multicolumn{2}{c}{DukeMTMC-ReID}\\
\cline{2-5}
&mAP&rank-1&mAP&rank-1\\
\hline
0.0&84.4&93.2& 73.8&85.6\\
0.2&84.2&93.5&73.9&86.2\\
0.4&83.8&93.0&73.6&85.5\\
0.6&83.1&92.8&73.8&\textbf{86.9}\\
0.8&84.1&93.3&73.8&86.0 \\
1.0&\textbf{84.5}&\textbf{93.6}&\textbf{74.2}&86.0\\
1.5&83.1&93.2&73.0&85.2\\
\hline
\end{tabular}
}
\end{table}

{\bf{Effectiveness of the camera domain alignment. }} To explore the effect of the alignment module, we visualized the feature distribution in early training epochs. As shown in Fig~\ref{img5}, feature distribution without the camera domain alignment module is somewhat biased towards camera labels. By comparison, features from different cameras show more similar distribution after introducing the alignment module, which effectively alleviates the distribution discrepancy and handles the intra-identity variance caused by the camera bias. We also evaluate the availability of the camera domain alignment module in boosting performance. With results presented in Table~\ref{table3}. It validates the necessity of the component.

{\bf{Effectiveness of probability decaying strategy. }}Probability decaying is developed to gradually enhance the complexity of training samples and imrove the robustness of the Re-ID model. To investigate the effectiveness of decaying the probability, we implement ablation study on Market1501 and DukeMTMC-ReID. Decay strategies include linear, polynomial, exponential and cosine decaying. All results are summarized in Table~\ref{table4}. Compared with the no-decaying model, the models with probability decaying do not always have better performance. The linear, exponential and cosine decaying benefit the final performance on Market1501. And only the cosine decaying can enhance performance on DukeMTMC-ReID. Our model achieves optimal performance when employing the cosine decay strategy, validating its effectiveness.

\begin{figure}[!t]
\centering
\includegraphics[width = \linewidth]{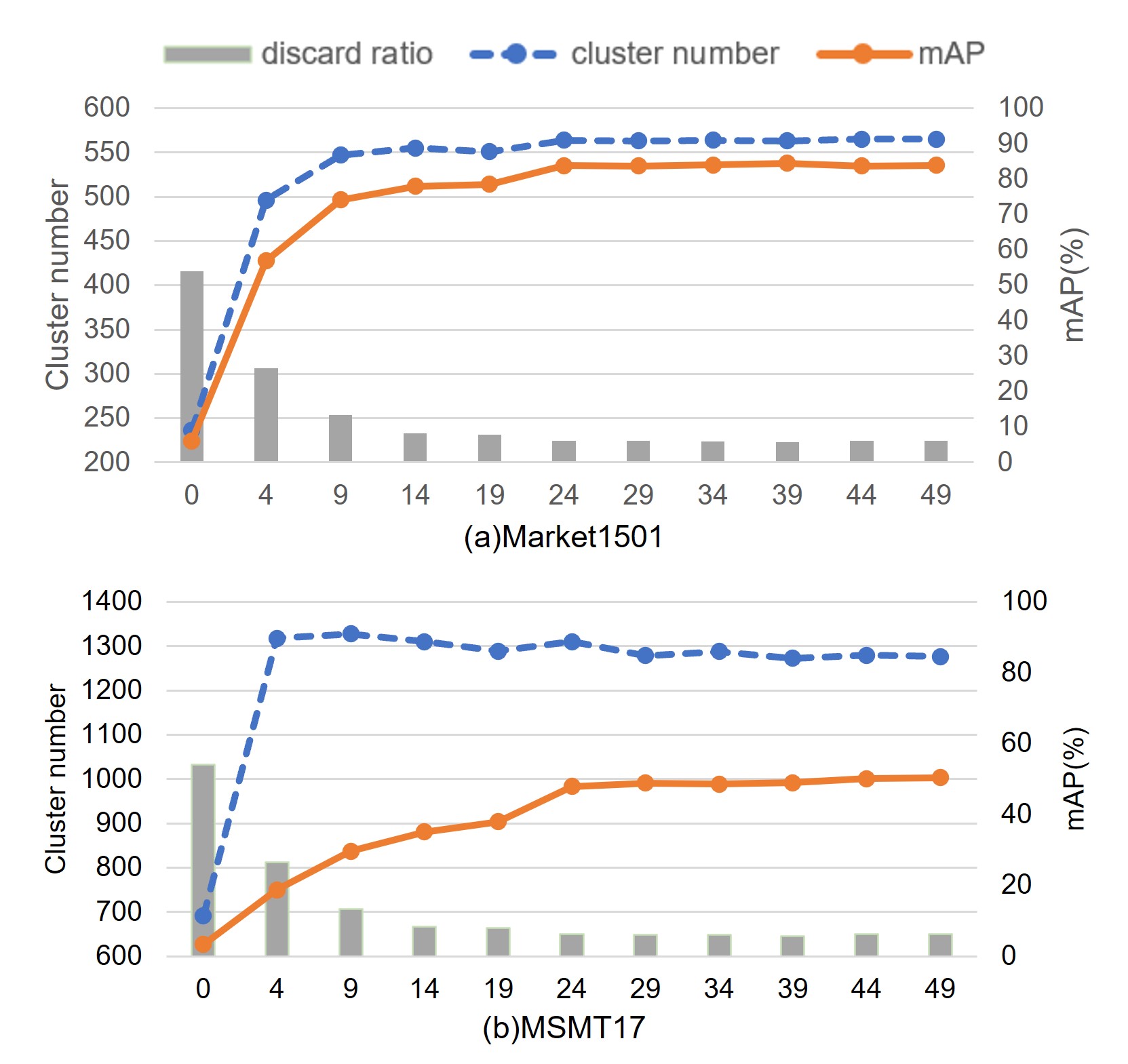}
\caption{The variation of the samples discarding ratio, cluster number and performance with training epochs.}
\label{img6}
\end{figure}

\subsection{Parameters analysis}
We analyze the impact of hyper-parameter $\beta$ to balance the inter-camera contrastive loss and domain classification loss (Eq.\eqref{e9}). Specifically, we keep the other parameter fixed and tune the parameter $\beta$ value. The experiment results are showed in Table \ref{table5}. When $\beta$ = 1.0, our proposed CALR achieves the highest rank-10 on Market1501. when $\beta$ = 0.6, CALR achieves the best performance on DukeMTMC-reID. and when $\beta$ = 1.0, CALR obtains the highest mAP.

In addition, we visualize the samples discarding ratio, cluster numbers and mAP with different training epochs on Market1501 and MSMT17 in Fig.~\ref{img6}. In this experiment, we don't use any probability decaying strategy. For the first epoch, we discard more than half of the samples for label refinement. It is intuitive because there are many global clustering errors in the initial state. After the label refinement, the remaining samples are more simple and reliable. We observe that mAP ascends rapidly at an early training epoch and then slowly converges. And discard ratio gradually decreases and then stabilizes with model training. Compared with other methods, our proposed CALR converges faster and obtains a considerable performance at about 30 epochs, which effectively saves training time. 

\begin{figure}[!t]
\centering
\includegraphics[width = \linewidth]{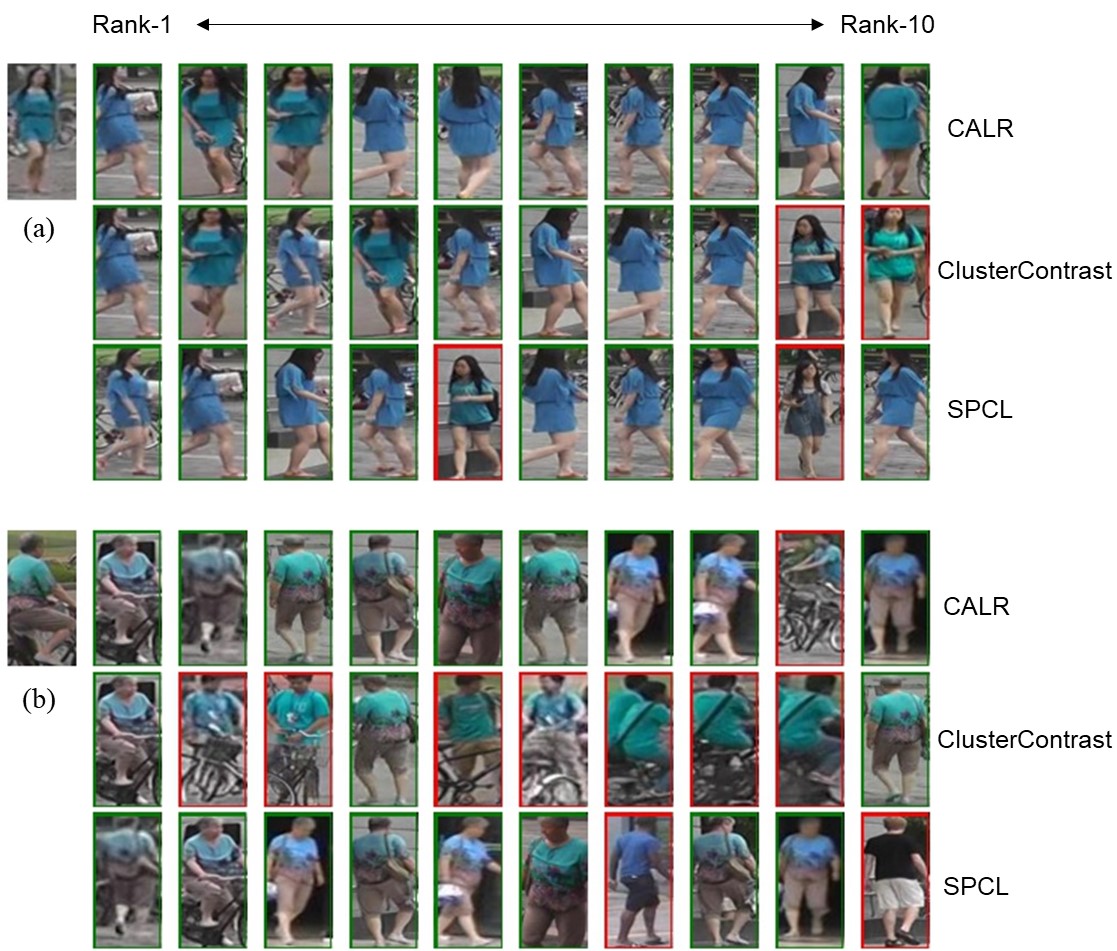}
\caption{The visual comparison of retrieval results among our proposed CALR, ClusterContrast~\cite{dai2022cluster} and SPCL~\cite{ge2020self} on Market1501. The correct samples are marked with green bounding boxes, while the wrong samples are marked with red bounding boxes.}
\label{img7}
\end{figure}

\subsection{Qualitative analysis}
We illustrate the retrieval results of our proposed CALR and the two popular methods ClusterContrast~\cite{dai2022cluster} and SPCL~\cite{ge2020self} on Market1501. Fig.~\ref{img7} show the visual comparison of retrieval images from two pedestrian queries. The matched images are sorted from left to right based on their Euclidean distance. Even if the pedestrian appearance of the query in Fig.~\ref{img7} (a) is not occluded, the clothing appears in different colors under different cameras, making it easier to match the wrong person. The second and third rows both retrieve incorrect pedestrian images that have very similar clothing to the query. However, the retrieval results of CALR are correct, demonstrating that CALR is robust to the large intra-identity appearance variance caused by camera settings. As for the second query in Fig \ref{img7} (b), the appearance of the pedestrian is occluded due to part of the body is not in the camera shooting area. And the pedestrian pose to ride a bike is not a regular pose either. ClusterContrast tends to retrieve the images of similar cyclists on the same camera. In comparison, our method better matches bicycling and walking images of the same pedestrian, validating the effectiveness of CALR in noisy image retrieving.  

\section{Conclusion} \label{sec5}
In this study, we introduce a novel Camera-Aware Label Refinement (CALR) framework designed to address the challenges of unsupervised person Re-ID by clustering intra-camera similarity. Our approach aims to mitigate the feature distribution bias and the inherent label noise caused by the camera bias. Specifically, we employ a camera alignment module to mitigate the camera distribution discrepancy. In terms of inherent label noise, we argue that feature distribution in a single camera has less intra-identity variance and intra-camera similarity relies on the appearance of the person instead of environmental factors. Therefore, performing intra-camera training obtains reliable local labels. We further design a criterion to select pivots to refine global clusters with local results. The refined global pseudo labels are adopted to compute the intra-camera contrastive loss for model updating. Extensive experiments on Market1501, DukeMTMC-reID, MSMT17 and Veri-776 demonstrate that our proposed CALR effectively improves the re-identification performance.

\bibliographystyle{IEEEtran}
\bibliography{string}{}     

%\newpage

\begin{IEEEbiography}[{\includegraphics[width=1in,height=1.25in, clip,keepaspectratio]{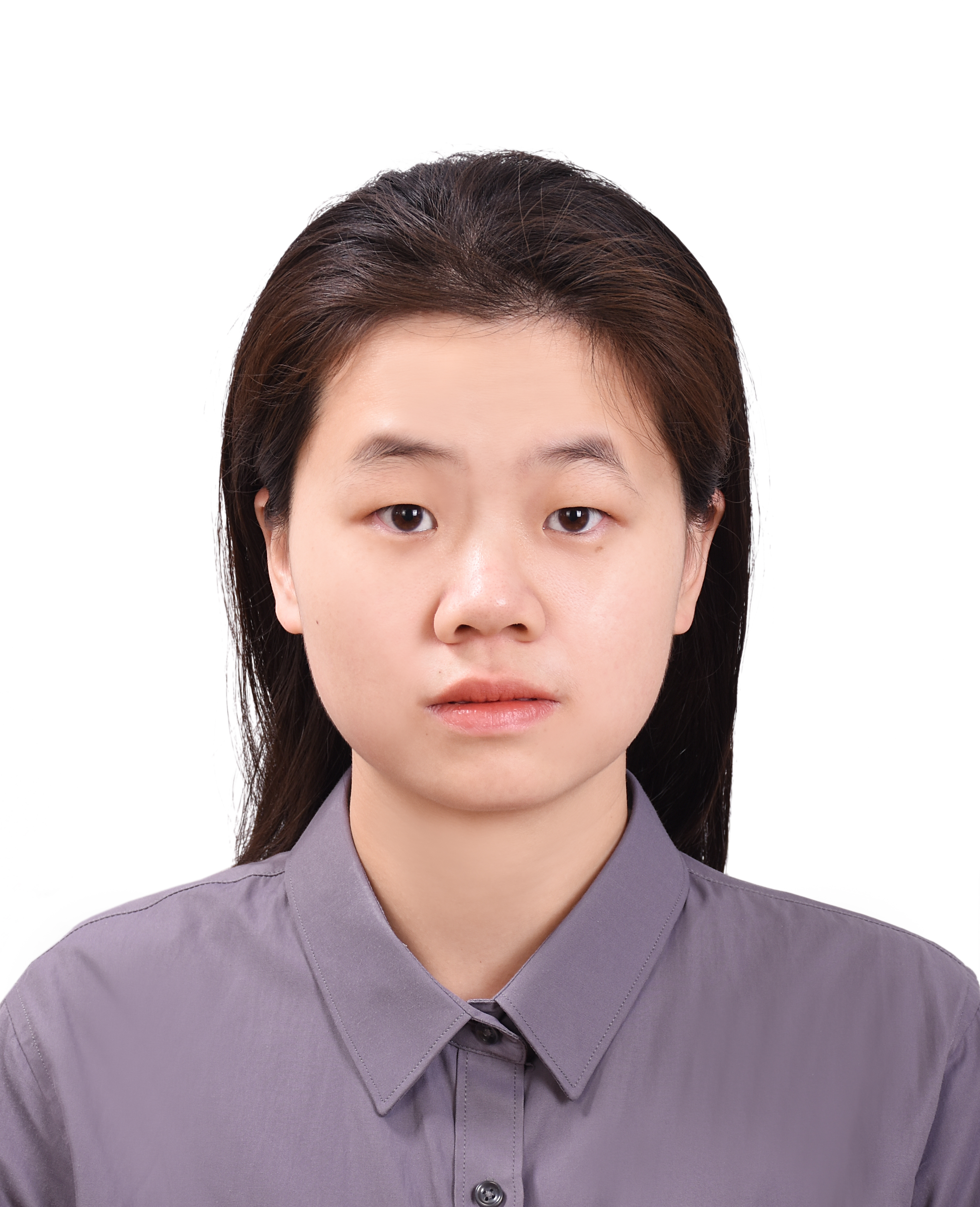}}]{Pengna Li} is currently pursuing Ph.D. degree with the Institute of Artificial Intelligence and Robotics, Xi’an Jiaotong University, Xi'an, China. She received her bachelors from the Institute of Automation, Xi’an Jiaotong University in 2021. Her research interests include computer vision, person re-identification and visual navigation.
\end{IEEEbiography}
\begin{IEEEbiography}[{\includegraphics[width=1in,height=1.25in, clip,keepaspectratio]{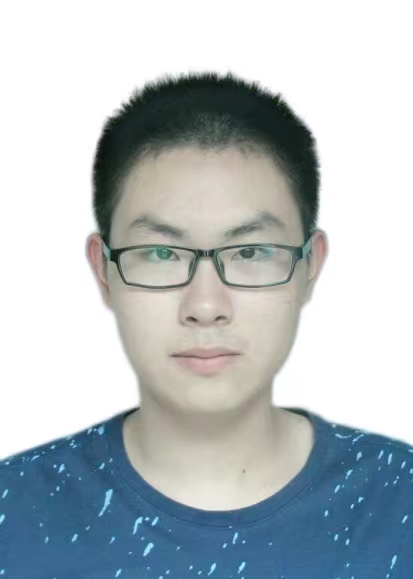}}]{Kangyi Wu} received the B.S. degree in control science and engineering from the Xi’an Jiaotong University, Xi’an, China, in 2022. He is currently pursuing Ph.D. degree with the Institute of Artificial Intelligence and Robotics, Xi’an Jiaotong University, Xi'an, China. His research interests include computer vision, deep learning, person re-identification and talking face generation.
\end{IEEEbiography}

\begin{IEEEbiography}[{\includegraphics[width=1in,height=1.25in,clip,keepaspectratio]{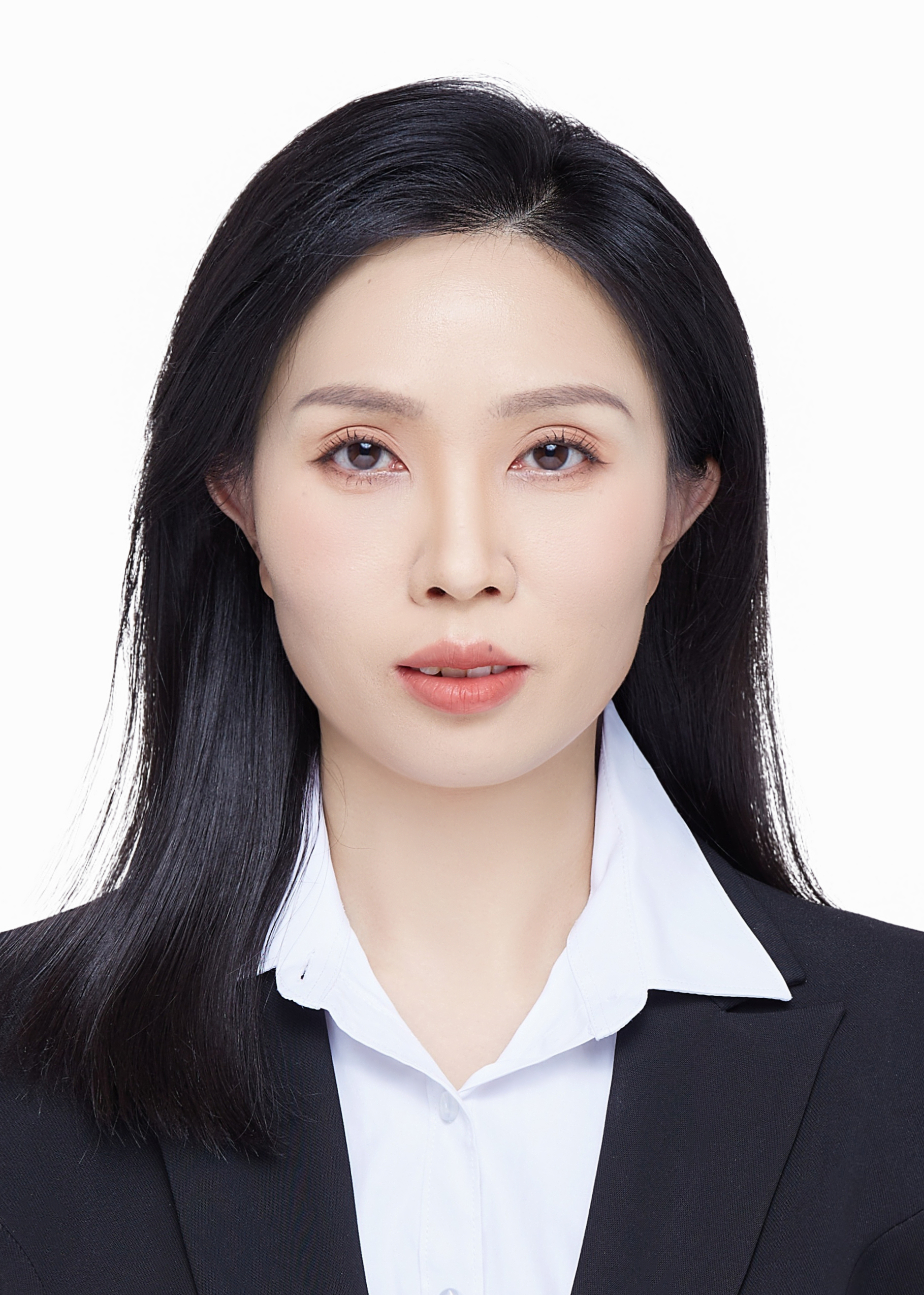}}]{Wenli Huang} received the Ph.D. degree from Xi’an Jiaotong University, Xi'an, China, in 2023. She is currently a Lecturer with the Ningbo University of Technology. Her research interests include deep learning and computer vision, with a focus on image inpainting, image restoration, image generation and person re-identification.
\end{IEEEbiography}

\begin{IEEEbiography}[{\includegraphics[width=1in,height=1.25in,clip,keepaspectratio]{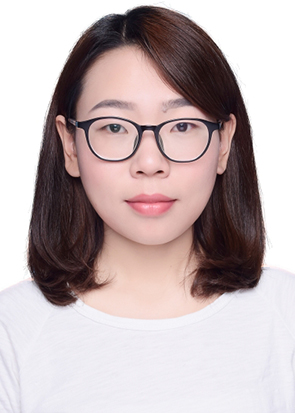}}]{Yang Wu} is currently pursuing the Ph.D. degree in the Institute of Artificial Intelligence and Robotics at Xi’an Jiaotong University. Her research interests include computer vision, image restoration, and graph representation learning.
\end{IEEEbiography}

\begin{IEEEbiography}[{\includegraphics[width=1in,height=1.25in,clip,keepaspectratio]{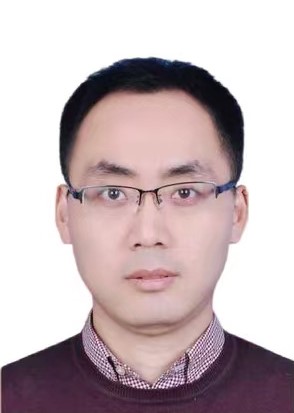}}]{Sanping Zhou} received his Ph.D. degree from Xi’an Jiaotong University, Xi’an, China, in 2020. He worked as a visiting researcher in the Human Sensing Lab at Carnegie Mellon University for one year's study in 2018. He is currently an Assistant Professor with the Institute of Artificial Intelligence and Robotics at Xi’an Jiaotong University. His research interests include machine learning and computer vision, with a focus on deep learning based algorithms in terms of object detection, multi-target tracking, person re-identification,  salient object detection, multi-task learning and meta-learning.
\end{IEEEbiography}

\begin{IEEEbiography}[{\includegraphics[width=1in,height=1.25in,clip,keepaspectratio]{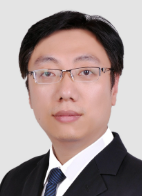}}]{Jinjun Wang} received the Ph.D. degree from Nanyang Technological University, Singapore in 2008. From 2006 to 2013, he worked in Silicon Valley, USA for leading research institutes including NEC Laboratories America, Inc. and Epson Research and Development, Inc. as a research scientist and senior research scientist. He is currently a Professor with Xi’an Jiaotong University. He has over 70 high-quality academic papers in prestigious international journals and conferences, including IEEE Trans. Multimedia, IEEE Trans. Intelligence Transportation Systems, CVPR, ACM MM. His research interests include computer vision, pattern recognition, multimedia computing and machine learning.
\end{IEEEbiography}

\vspace{11pt}

\vfill

\end{document}